\documentclass{article}
\usepackage{spconf,amsmath,graphicx,hyperref}
\usepackage{amsmath,graphicx,url,times}
\usepackage{amssymb}
\usepackage{color}
\usepackage{bm}
\title{Large Language Models Implicitly Learn to See and Hear\\  Just By Reading}
\name{Prateek Verma \qquad Mert Pilanci}
\address{Department of Electrical Engineering \\
Stanford University \\
Stanford, CA 94305, USA}

\usepackage{cuted}
\usepackage{capt-of}
\usepackage[symbol]{footmisc}

\begin{document}
%\ninept
%
\maketitle
%\begin{sloppy}
%\begin{strip}\centering
%\includegraphics[width=0.9\linewidth,height=7.02cm,keepaspectratio]{fig2.png}
%\captionof{figure}{Overview of our proposed method in the paper. We replace ViT/Audio-Transformer encoder weights with weights of text-LLM pretrained on next text token prediction, enabling them to see and hear using learned circuits just by reading text. 
%\label{fig:feature-graphic}}
%\end{strip}
%\end{sloppy}

\begin{abstract}
This paper presents a fascinating find: By training an auto-regressive LLM model on text tokens, the text model inherently develops internally an ability to understand images and audio, thereby developing the ability to see and hear just by reading. Popular audio and visual LLM models fine-tune text LLM models to give text output conditioned on images and audio embeddings. On the other hand, our architecture takes in patches of images, audio waveforms or tokens as input. It gives us the embeddings or category labels typical of a classification pipeline. We show the generality of text weights in aiding audio classification for datasets FSD-50K and GTZAN. Further, we show our framework working for image classification on CIFAR-10 and Fashion-MNIST, as well on image patches. This pushes the notion of text-LLMs learning powerful internal circuits that can be utilized by activating necessary connections for various applications rather than training similar models from scratch every single time.
\end{abstract}
\section{Introduction}
\label{sec:intro}
\vspace{-0.2cm}
Large Language Models (LLMs) have profoundly impacted AI, enabling advances unimaginable just a few years ago, such as winning medals in the International Math Olympiad \cite{trinh2024solving}. Their success has driven diverse domains—natural language processing \cite{brown2020language}, acoustic tokens \cite{borsos2023audiolm,verma2020framework}, vision \cite{yan2021videogpt}, robotics \cite{brohan2023rt2}, proteins \cite{madani2020progen}, raw waveforms \cite{verma2021generative}—to converge on a GPT-like pipeline. Here, the modality is tokenized, and a GPT model predicts the next token, with reconstruction applied when needed for the target domain. Post-training methods such as test-time scaling \cite{muennighoff2025s1} further enhance LLM performance. The primary motivation for this work comes from three papers. First, the Audio Spectrogram Transformer (AST) by Glass et al. \cite{gong21b_interspeech} showed that Vision Transformer (ViT) weights \cite{dosovitskiy2020image}, trained only on ImageNet, could generalize to audio spectrograms. This is intuitive, as spectrograms share structural similarity with images and have historically used similar neural architectures, e.g., CNNs \cite{hershey2017cnn}. Second, AudioPALM \cite{rubenstein2023audiopalm} leveraged the strong link between spoken language and text. Since text compresses prosody, energy, and style, AudioPALM mapped acoustic tokens to Byte Pair Encoded text tokens and vice versa, while keeping the text LLM backbone frozen, enabling speech-to-text embedding transfer. Third, frozen Transformers were shown to act as Universal Compute Engines (UCE) \cite{lu2022frozen}, without leveraging finetuning methods such as LORA\begin{figure}[t]
  \centering
  \centerline{\includegraphics[width=\columnwidth]{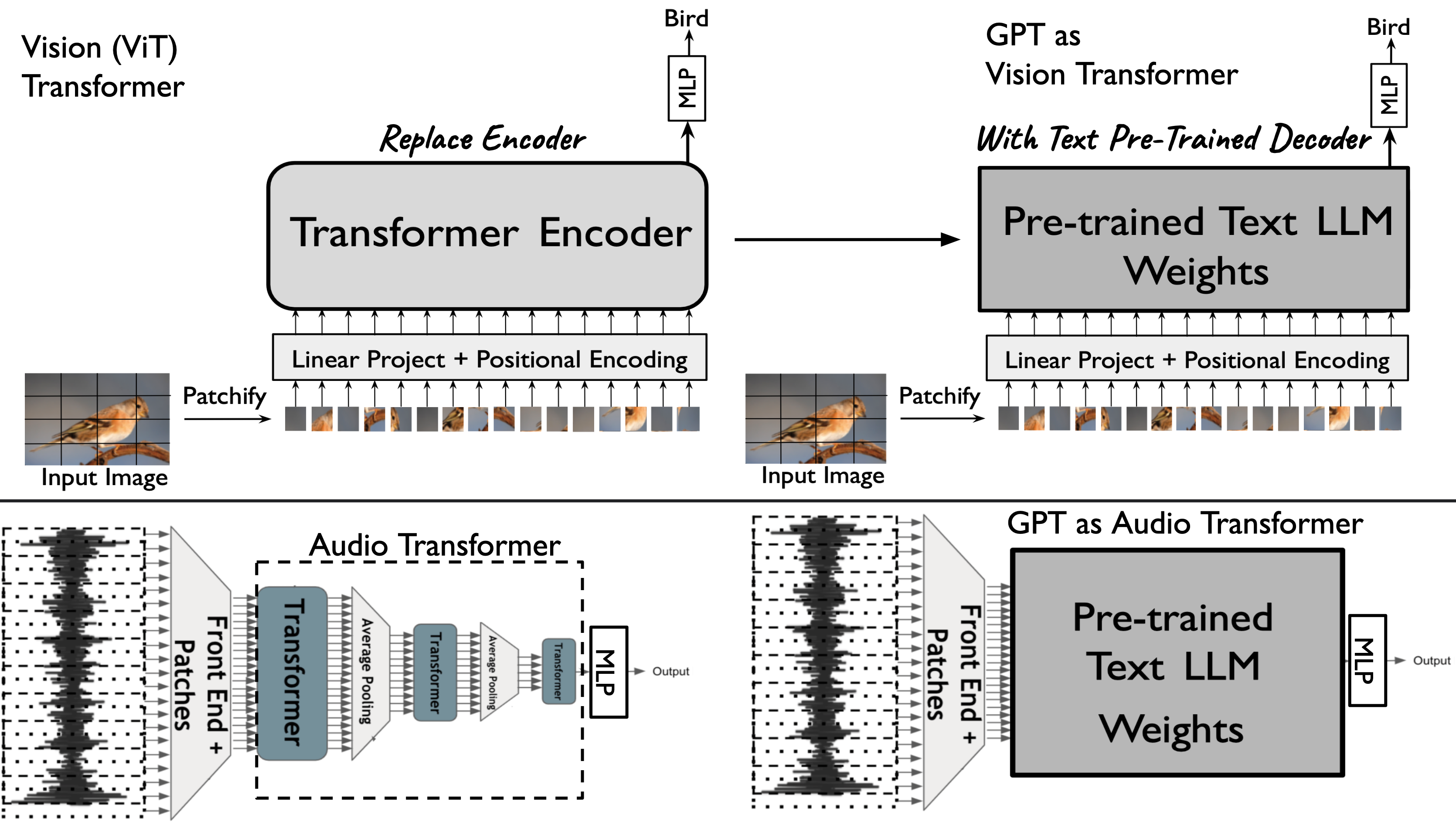}}
  \caption{Overview of our proposed method. We replace ViT/Audio-Transformer encoder weights with weights of text-LLM pretrained on next text token prediction, enabling them to see and hear using learned circuits just by reading. }
  \vspace{-0.55cm}
  \label{fig:results}
\end{figure}. In this work, we extend prior approaches by showing that finetuning existing text pretrained LLMs can surpasses frozen architectures, achieving performance comparable to models trained from scratch. This result generalizes to audio understanding, which has not been shown before. While GPT-style models are pretrained solely on billions of text tokens with no connection to images or audio, their attention mechanism captures complex dependencies useful across modalities. We demonstrate that the idea of a unified speech–text modality in AudioPALM extends to domains unrelated to text. By finetuning, we activate internal circuits, enabling a pre-trained text LLM to replace standard classification pipelines and outperform frozen LLM approaches on UCE. Moreover, \cite{prakash2024finetuning} showed that finetuning uncovers powerful internal structures: for entity tracking, pretrained text LLMs fine-tuned on structured domains such as math achieved substantial gains, outperforming the original LLM even on language data. The contributions of this work are: i) Encoder-free learning: We show that text-LLM weights trained on next-token prediction can classify image or audio patches without separate encoders. A text-LLM directly acts as an encoder, mapping each chunk to an embedding typical of classification pipelines. ii) We surpass frozen GPT-weight architectures such as AudioPALM and UCE by applying PEFT methods particularly LORA \cite{hu2021lora}, since finetuning intuitively enhances performance. iii) Scaling: Increasing base text only LLM size keeping other components fixed, improves classification across all datasets and modality including images and audio. Our method has the potential to improve almost any classification pipeline.
\section{Dataset}
\label{sec:data}
For this paper, we demonstrate our method of utilizing text LLM weights pretrained on next-token prediction across two classification modalities: visual recognition and audio. For images, we use CIFAR-10 and Fashion-MNIST, two widely used academic benchmarks. CIFAR-10 contains 10 categories with 50k training and 10k test images, while Fashion-MNIST is a 10-class grayscale dataset. For audio, we benchmark on FSD-50K \cite{fonseca2020fsd50k}, a popular dataset with multi-class labels per audio clip, similar to AudioSet. We chose FSD-50K because of its standardized training–testing pipeline, including augmentations and context lengths, ensuring fair comparisons. Finally, to highlight the method’s versatility, we model coarse acoustic tokens for classification, aligning with approaches such as AudioPALM on the GTZAN dataset \cite{tzanetakis_essl_cook_2001}, which classifies 1s audio chunks into 10 music genres using coarse representations.

\section{Methodology}
\label{sec:method}
\subsection{Difference with AudioLLM} We categorize AudioLLMs into two main architectures. While parallels exist in vision, we restrict discussion to audio. The first class consists of decoder-only models that operate on audio token sequences generated by tokenizers such as EnCodec \cite{defossez2022high}, mirroring autoregressive language models like GPT \cite{brown2020language}, with next-token prediction as the training objective. Google initiated this approach with discrete representation learning \cite{van2017neural}, later extending it to Jukebox \cite{dhariwal2020jukebox}, VALL-E \cite{wang2023neural}. Without encoders, these models represent canonical decoder-only AudioLLMs.
The second class maps audio to text using encoder-decoder setups. Audio is encoded into embeddings or tokens, which condition fine-tuned text LLMs (e.g., GPT, LLaMA) for autoregressive generation. These models target tasks such as captioning and audio QA, predicting text tokens from audio prompts. Architecturally, they resemble vision-language models like Flamingo \cite{alayrac2022flamingo}, BLIP\cite{li2022blip} and LLaVA \cite{liu2023visual} which connect separate modality encoder to a decoder; AudioFlamingo \cite{kong2024audio} directly extends this to audio. Unlike decoder-only AudioLLMs, these are task-specific adaptations of text LLMs. Our architecture diverges by directly fine-tuning pretrained text-only LLM weights for non-text modalities. As shown in Fig. 2 (red), we adapt these weights as classifiers/embedding extractors, enabling representation learning across audio and images. Unlike LTU-AS \cite{gong2023joint} or Qwen-Audio \cite{chu2023qwen}, which condition on audio encoder embeddings, we fine-tune the LLM itself. Similarly, while Audio Transformer \cite{verma2021audio} (Fig. 1) projects embeddings to class labels via MLPs, we hypothesize that Transformer \cite{vaswani2017attention} components—attention and feed-forward layers—trained on text prediction are modality-agnostic. Empirically, minimal fine-tuning outperforms models trained from scratch, and performance scales with GPT backbone size, consistent with scaling laws \cite{kaplan2020scaling} in text-only LLMs.
\begin{figure}[t]
  \centering
  \centerline{\includegraphics[width=\columnwidth]{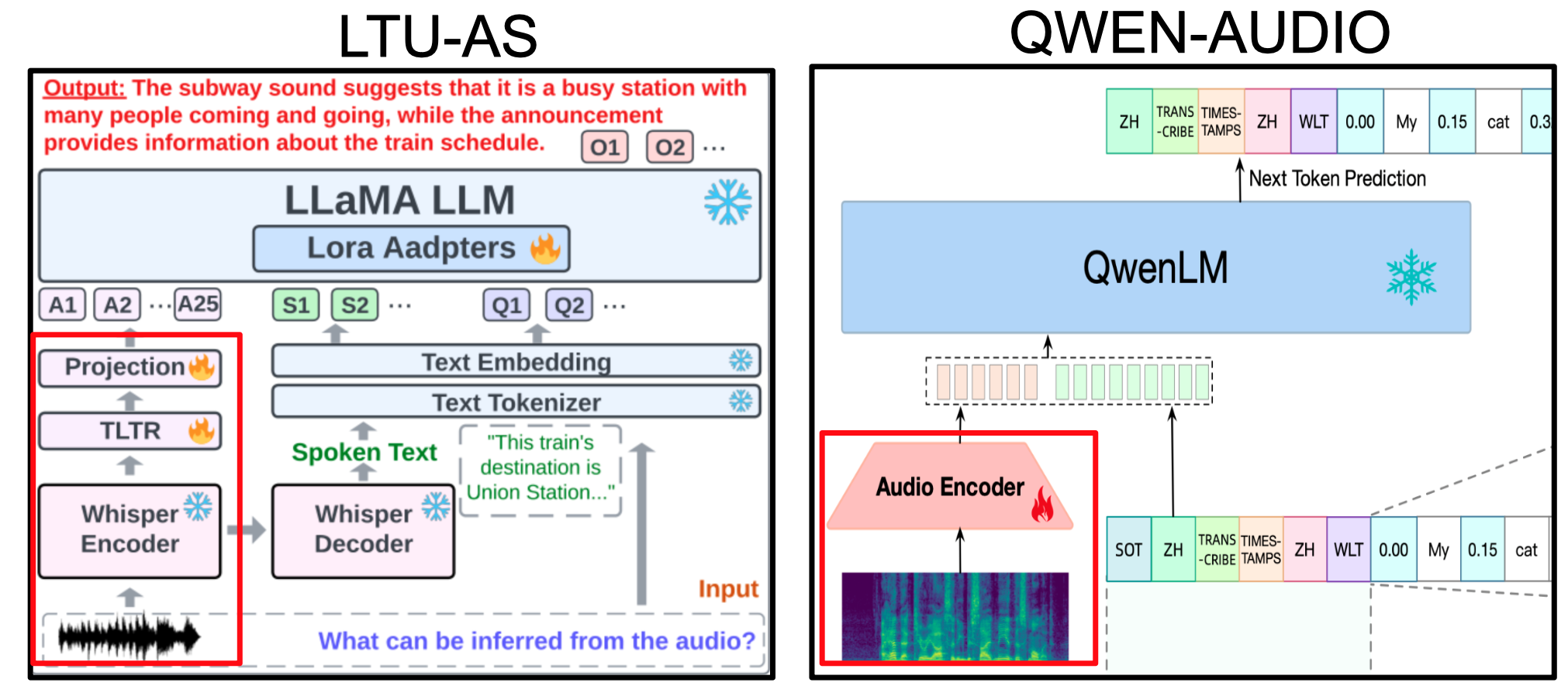}}
  \caption{Two popular AudioLLM architectures are shown, with figures from LTU-AS \cite{gong2023joint} and Qwen-Audio \cite{chu2023qwen}. The red block marks the component replaced. We learn an audio encoder—a classifier or pretrained model that maps audio representations to embedding vectors via a pretrained text-LLM.}
  \label{fig:results}
\vspace{-0.4cm}
\end{figure}

\vspace{-0.4cm}
\subsection{Training Recipe}
Given an input data modality of interest, $\mathbf{X} \in \mathbb{R}^{H \times W \times C}$, we divide it into non-overlapping patches of size $P \times P$, giving $N = H \cdot W / P^2$ patches. Each patch is flattened into a vector, forming patch tokens $\mathbf{X}_p \in \mathbb{R}^{N \times (P^2 \cdot C)}$. A linear projection is applied to obtain patch embeddings $\mathbf{E} = \mathbf{X}_p \mathbf{W}_e + \mathbf{b}_e$, where $\mathbf{W}_e \in \mathbb{R}^{(P^2 \cdot C) \times D}$ and $\mathbf{E} \in \mathbb{R}^{N \times D}$. To retain spatial information, positional embeddings are added: $\mathbf{X_e} = \mathbf{E} + \mathbf{P}$, with $\mathbf{P} \in \mathbb{R}^{N \times D}$. The input embedding $\mathbf{X_e}$ is passed to a Transformer encoder, a computation block applicable to multiple modalities with varying dimensions ($C, H, W$). For instance, $C=3$ for color images like CIFAR-10, $H=1$ for 1-D audio waveforms, and $C=1$ for grayscale images or audio spectrograms. For audio, the patch size must be selected carefully to respect the quasi-stationary properties of waveforms, enabling the linear projection to learn effective representations. This positionally-encoded patch embedding, $X_e \in \mathbb{R}^{(N+1) \times D}$, serves as the initial input $X^{(0)}$ to a stack of $L$ identical layers. Each layer comprises a multi-head self-attention block and a feed-forward network, utilizing residual connections and layer normalization. The output of the final layer is $X^{(L)}$. This core architecture is shared by AST, Audio Transformers, and ViT, differing primarily in learned parameters and task-specific labels. The structure is analogous to LLM architectures, which mainly differ by employing causal attention masks in the Transformer stack. \textbf{This paper explores the question: Can we reuse the same set of learnable parameters for the next token prediction for problems that have no connection to that of text or output text tokens, e.g. using text-GPT weights to image and audio classifier?}
Intuitively, prior works have succeeded in related domains. AudioPALM leveraged shared speech-text spaces to build foundational speech architectures by mapping text tokens to speech tokens. Similarly, AST utilized ViT weights for audio classification on spectrogram inputs, as 2-D mel spectrograms can be visually interpreted by trained researchers. We extend this approach by employing GPT weights trained on internet-scale data, capturing universal dependencies to accurately predict the next token. These weights are repurposed for image and audio classification. Following AudioPALM/AST, we remove the pipeline mapping GPT-2 text tokens and instead convert input image or waveform representations $\mathbf{X} \in \mathbb{R}^{H \times W \times C}$ into patches, linearly mapping them to embeddings $\mathbf{X_e}$ using learned matrices $\mathbf{W_e}$ and $\mathbf{b_e}$, yielding $\mathbf{E_n} = \mathbf{X_p} \mathbf{W_{ne}} + \mathbf{b_{ne}}$. These embeddings are combined with the original learned positional embeddings of GPT-2, which remain unchanged, ensuring compatibility with the Transformer stack. Consequently, we only relearn $\mathbf{W_{ne}}, \mathbf{b_{ne}}$ to map the new modality to the space expected by the pretrained text decoder layers, akin to AudioPALM but applied to continuous inputs such as image pixels or audio waveforms.
While AudioPALM/AST freeze the base model weights, we incorporate parameter-efficient fine-tuning, e.g., LoRA, to adjust the model weights as needed. We allow parameter-efficient fine-tuning like LoRA to tweak the weights. We fix low rank $r$ and $\alpha$ at 8 throughout the paper, without performing a dataset- or architecture-specific grid search. While tuning these could further improve performance, our results are already strong. The only modification is taking the last token from the GPT-2 backbone output and applying an MLP to classify embeddings learned from GPT. Intuitively, this maps the target modality to the GPT embedding space and decodes it via the last layer. This generic pipeline can replace \textbf{any} classification setup, including vision (ViT) or audio (Audio Transformer), which learn embeddings and Transformer weights from scratch.
The trainable parameters are limited to the added PEFT layers and input/output projection matrices. We demonstrate that using a pretrained LLM backbone achieves near-comparable performance with minimal learnable parameters. This approach emphasizes improving front-end architectures rather than scaling parameters or data, as in Whisper \cite{radford22whisper} with sub-optimal mel-spectrogram front-ends \cite{verma2016frequency}. For audio, better front-ends \cite{zeghidour2021leaf,verma2023content} could further boost performance, in contrast to massive backbones like One-Peace \cite{wang2023one}, infeasible for academia. 

\section{Results And Discussion}\label{sec:results} To demonstrate our method, we evaluate on representative audio and image datasets, comparing against fine-tuned baselines such as ViT and against models of similar size trained from scratch. We further contrast frozen vs. tunable LLM backbones, showing that PEFT applied to text-only LLMs yields competitive results on both audio and images. 
\vspace{-0.4cm}\begin{figure}[t]
  \centering  \centerline{\includegraphics[width=\columnwidth]{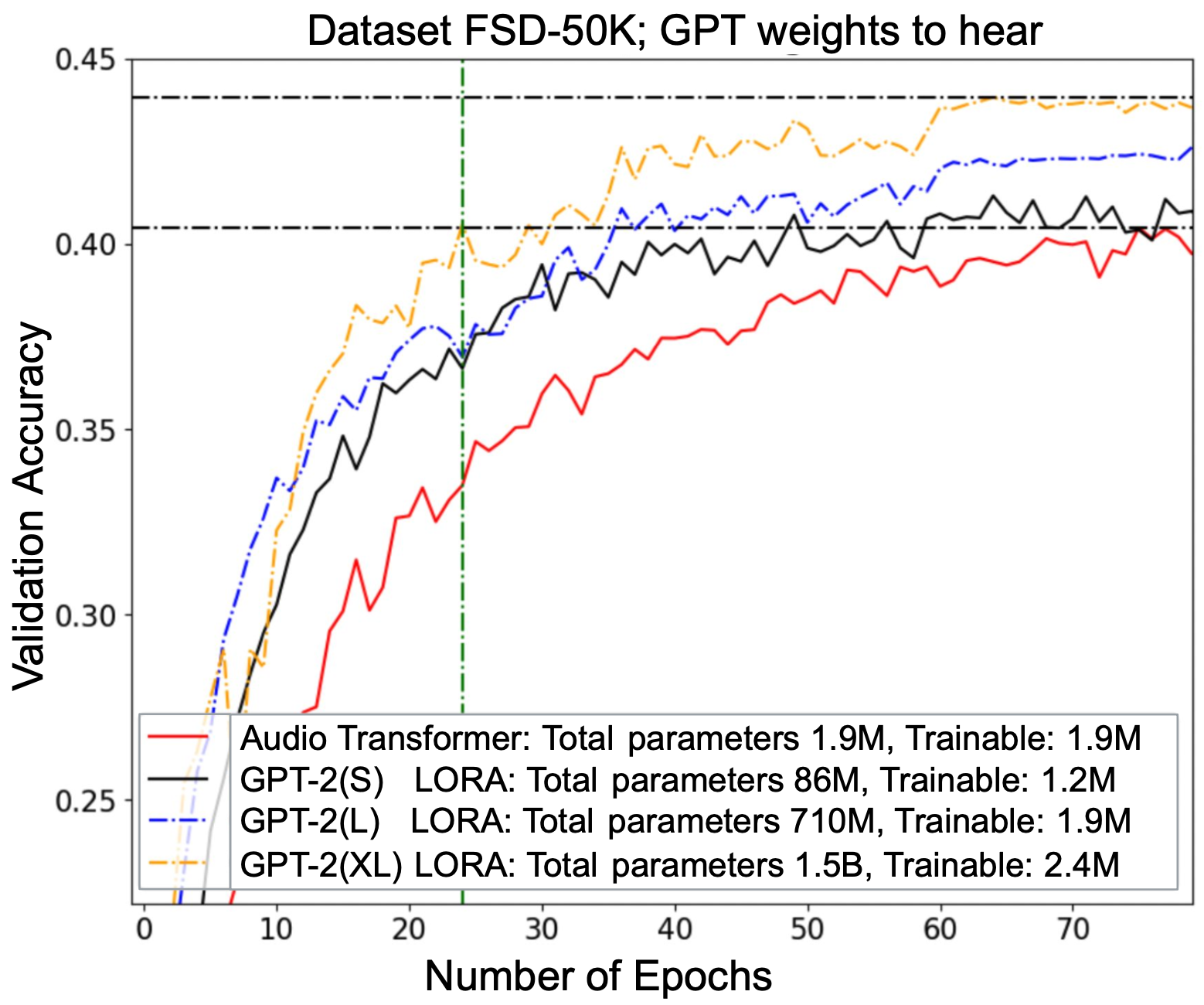}}
  \caption{Results on FSD-50K show performance improvement with scale. All four models share a learnable front-end that patches the waveform and maps it to the first Transformer layer. Audio Transformer trains all parameters from scratch, while the others use GPT-2 weights pretrained on next-token prediction with LoRA tuning across different model sizes.}
  \label{fig:results}
\vspace{-0.4cm}
\end{figure} \subsection{Results on Images and Audio} We follow a ViT-style recipe for evaluation, where all architectures share the same patch embedding and positional encoding pipeline, ensuring fair comparison. The difference lies in subsequent processing: our method either keeps the text-LLM frozen, fine-tuning nothing, or applies PEFT via LoRA to update only a small parameter subset. We also compare against models trained from scratch on CIFAR-10 and fully fine-tuned ViT on Fashion-MNIST. Across datasets, our approach is comparable to scratch-trained/fine-tuned ViTs.
\begin{table}[h!]
\centering
\begin{tabular}{|l|l|l|c|}
\hline
\textbf{Model Config} & \textbf{Backbone} & \textbf{\#params} & \textbf{Accu} \\
\hline
GPT(S) LORA & GPT-87M & 0.64M  & 76.6 \% \\
GPT(M) LORA & GPT-340M & 1.4M &  78.4 \% \\
GPT(S)\cite{lu2021pretrained} & GPT -87M & Frozen & 72.1\%\\ 
ViT \cite{chh2021vit} & Scratch  & 86M & 80.1\% \\
\hline
\end{tabular}
\caption{Model variants on CIFAR-10: all models train patch embeddings from scratch. The Transformer stack differs as: (i) pretrained GPT-2 weights with LoRA, (ii) frozen GPT-2 weights, or (iii) ViT trained from scratch with random init.}
\label{tab:fashion-mnist-results}
\vspace{-0.6cm}
\end{table}
\subsection{Frozen Backbone vs LoRA based finetuning} Following AudioPalm \cite{rubenstein2023audiopalm} and UCE \cite{lu2022frozen}, we evaluate our model on CIFAR-10 with the base LLM frozen. As shown in Table 1, our approach significantly outperforms the frozen setup. Intuitively, fine-tuning even a small subset of parameters improves generalization, consistent with prior work \cite{hu2021lora}. Hence, for subsequent experiments on scale, audio representations, and training setups (including models trained from scratch), we focus on LoRA-tuned architectures.
\begin{table}[h!]
\centering
\begin{tabular}{|l|l|l|c|}
\hline
\textbf{Model Variant} & \textbf{Backbone} & \textbf{\#params} & \textbf{Accuracy} \\
\hline
GPT(S) & GPT - 87M & 0.6M  & 91.6\% \\
GPT(M) & GPT - 340M & 1.4M & 92.4\% \\
ViT \cite{bbouzidi2024convolutional} & ViT-Finetuned & 86M & 95.5\% \\
\hline
\end{tabular}
\caption{Different model variants on the Fashion-MNIST dataset. All models train patch embeddings from scratch, with only difference being in the Transformer stack, with using i) pretrained GPT-2 weights from next text token prediction followed by LORA ii) Fine-tuned ViT weight trained on images}
\label{tab:fashion-mnist-results}
\vspace{-0.6cm}
\end{table}
\subsection{Effect of Scale} We observe consistent accuracy gains as the backbone text LLM scales from 87M to nearly one billion parameters, consistent with LLM scaling laws where larger models and data improve performance \cite{kaplan2020scaling}. These improvements arise despite keeping the front end and linear classifier fixed for both images and audio, suggesting that fine-tuning leverages richer functions and connections in larger models. On audio genres, accuracy improved by nearly 2\%, while on continuous audio representations in FSD-50K, billion-parameter models yielded close to 4\% absolute gains. As shown in Figure 3, larger text models not only achieve higher accuracy in audio but also converge faster, indicating that fine-tuning activates latent audio-relevant circuits in pretrained text LLMs. \begin{table}[h!]
\centering
\begin{tabular}{|l|l|l|c|}
\hline
\textbf{Model Variant} & \textbf{Backbone} & \textbf{\#params} & \textbf{Accuracy} \\
\hline
GTZAN Scratch &  \hspace{2.2em}--\hspace{1em} &2.3M & 63.1\% \\
GTZAN LORA & GPT(S) - 87M & 1.2M & 66.7\% \\
GTZAN LORA & GPT(S) - 340M & 1.9M & 68.9\% \\
\hline
\end{tabular}
\caption{Results on 1s ENCODEC tokens for different setups, where all models train the embedding matrix from scratch (AudioPALM) with difference in Transformer stack: (i) pretrained GPT-2 weights with LoRA (ii) trained from scratch.}
\vspace{-0.55cm}
\end{table}
\subsection{Results on Audio}
\begin{table}[h!]
\centering
\begin{tabular}{|l|l|l|c|}
\hline
\textbf{Model} & \textbf{Backbone} & \textbf{\#param} & \textbf{Top-5 accu} \\
\hline
Small & GPT - 87M (LORA) & 1.2M & 41.4\% \\
Large & GPT - 710M(LORA) & 1.9M & 42.7\% \\
XLarge & GPT - 1.5B(LORA) & 2.4M & 44.1\% \\
Scratch\cite{verma2021audio} & Audio Transformer & 1.9M & 40.4\% \\
\hline
\end{tabular}
\caption{Results on audio waveform for FSD-50K dataset. All models train patch embeddings from scratch, with only difference being in
the Transformer stack, with either using i) pretrained GPT-2 weights from next text token prediction followed by LORA ii) Training Audio Transformer from scratch.}
\vspace{-0.5cm}
\end{table} We evaluate on GTZAN and FSD-50K, using the coarsest ENCODEC token input for GTZAN and raw waveforms for FSD-50K. For GTZAN, the task maps acoustic tokens to an embedding matrix following Audio-PALM \cite{rubenstein2023audiopalm}, but with a Transformer decoder backbone tunable via LORA for classification. For FSD-50K, the front end follows \cite{verma2021audio}, with 4 convolutional sets (64 filters, kernel sizes 16–128) applied to 1s audio in 25ms patches. Filter outputs are collapsed via max-pooling and ReLU to form a 256-dim patch representation, projected to the Transformer input. The decoder’s final output is classified using a linear head with Huber loss \cite{verma2021audio}. Like image setup, only the Transformer backbone varies—trained from scratch or fine-tuned from text-GPT with LORA. Fig ~1 shows fine-tuned backbones outperform scratch-trained models with comparable parameters/larger LLMs providing gains.
\section{Conclusion And Future Work}
\label{sec:futurework} We show how weights of a text-based LLM can be reused for downstream classification tasks in image, audio and music domains, with no similarity between text tokens and the input patch representation of pixels or waveform. Inspired by AudioPalm, we learn lightweight embedding matrices that map non-textual inputs into the space of the first layer of a text LLM. Unlike prior VLLM/ALLM/BLIP approaches that feed embeddings into a frozen text model, we show LLM's internal circuitry alone can suffice for direct classification, obviating the need for a separate encoder. Our approach achieves competitive performance using only a small fraction of parameters required by standard architectures. Scaling up text backbone LLM boosts accuracy in other domains like audio/images, suggesting a broad impact across modalities by leveraging circuitry inside text-pre-trained LLMs. They can be reused instead of everytime training new Transformer based models. 

% -------------------------------------------------------------------------
\bibliographystyle{IEEEbib}
\bibliography{refs}

\end{document}